\documentclass[letterpaper, 10 pt, conference]{ieeeconf} 
\IEEEoverridecommandlockouts                              
\usepackage[usenames,dvipsnames,table,xcdraw]{xcolor}
\usepackage{color}
\usepackage[utf8]{inputenc}
\usepackage{graphicx}
\usepackage{multicol}
\usepackage{multirow}
\usepackage{array}
\usepackage{footmisc}
\usepackage{amssymb}
\usepackage[hidelinks]{hyperref}
\usepackage{amsmath}
\usepackage{amsfonts}
\usepackage{hyperref}
\usepackage{algorithm}
\usepackage{comment}
\usepackage{xspace}
\usepackage{tikz}
\usepackage{float}
\usetikzlibrary{math,shapes,calc,decorations.text}
\usepackage[caption=false, font=footnotesize]{subfig}
\usepackage{listings}
\usepackage{xcolor}
\usepackage{booktabs} 
\usepackage{algpseudocode}
\usepackage[backend=biber,
            url=false,
            isbn=false,
            doi=false,
            backref=false,
            style=ieee,
            natbib=true,
            mincitenames=1,
            maxcitenames=1,
            maxbibnames=10,
            citestyle=numeric-verb,
            sorting=none,
            block=none]{biblatex}

\bibliography{./references.bib}

\newcommand{\remark}[3]{\definecolor{#1}{hsb}{#2,1.0,0.7}{\color{#1}[#1: #3]}}

\newcommand{\todo}[1]{\remark{TODO}{0.5}{#1}}

\newcommand{\AlgName}{STITCH\xspace}

\usepackage{amsmath}

\title{\LARGE \bf
\AlgName: Augmented Dexterity for Suture Throws \\ Including Thread Coordination and Handoffs
}

\author{Kush Hari$^{1*}$, Hansoul Kim$^{1*}$, Will Panitch$^{1*}$, Kishore Srinivas$^{1}$, Vincent Schorp$^{1,2}$, \\
Karthik Dharmarajan$^{1}$, Shreya Ganti$^{1}$,
Tara Sadjadpour$^{1}$, Ken Goldberg$^{1}$
\thanks{The AUTOLab at the University of California, Berkeley (\url{automation.berkeley.edu}), 
{\tt\small goldberg@berkeley.edu}}
\thanks{* equal contribution}
\thanks{$^{1}$ AUTOLab at University of California, Berkeley} 
\thanks{$^{2}$ Autonomous Systems Lab, ETH Zurich}
}

\begin{document}

\maketitle
\thispagestyle{empty}
\pagestyle{empty}

\begin{abstract}
We present STITCH: an augmented dexterity pipeline that performs \underline{S}uture \underline{T}hrows \underline{I}ncluding \underline{T}hread \underline{C}oordination and \underline{H}andoffs. STITCH iteratively performs needle insertion, thread sweeping, needle extraction, suture cinching, needle handover, and needle pose correction with failure recovery policies. We introduce a novel visual 6D needle pose estimation framework using a stereo camera pair and new suturing motion primitives. We compare STITCH to baselines, including a proprioception-only and a policy without visual servoing. In physical experiments across 15 trials, STITCH achieves an average of 2.93 sutures without human intervention and 4.47 sutures with human intervention. See \url{https://sites.google.com/berkeley.edu/stitch} for code and supplemental materials. 

\end{abstract}

\section{Introduction}


Surgical-assist robots such as Intuitive's da Vinci series are used by surgeons to perform over 2 million  procedures annually to facilitate minimally-invasive (“keyhole”) surgery to reduce pain, blood loss, scarring, complications, and recovery time. It is now the gold standard for procedures involving the appendix, colon, gallbladder, prostate, and many others. These robots are very sophisticated but every movement is currently controlled  by human surgeons because surgery is extremely sensitive to errors – there are a vast number of rare but potentially dangerous edge conditions and the consequences of even a single failure can be fatal. So it may be a very long time before fully autonomous robots are sufficiently safe and reliable for the operating room. However, recent advances in AI are opening the door to augmenting surgeon skills when performing specific subtasks such as suturing, debridement, and resection. Goldberg \cite{goldberg_augmented_2023} proposed the term “Augmented Dexterity” to describe systems where surgical subtasks are controlled by the robot under the close supervision of the human surgeon who is ready to take over at a moment’s notice.  Augmented Dexterity has potential to elevate good surgeons to the level of the best surgeons, making surgery safer, faster, and more reliable.  In this paper we present new results on augmented dexterity for surgical suturing.



\begin{figure}[t]
    \centering
    \vspace{0.08in}\includegraphics[width=1.0\linewidth]{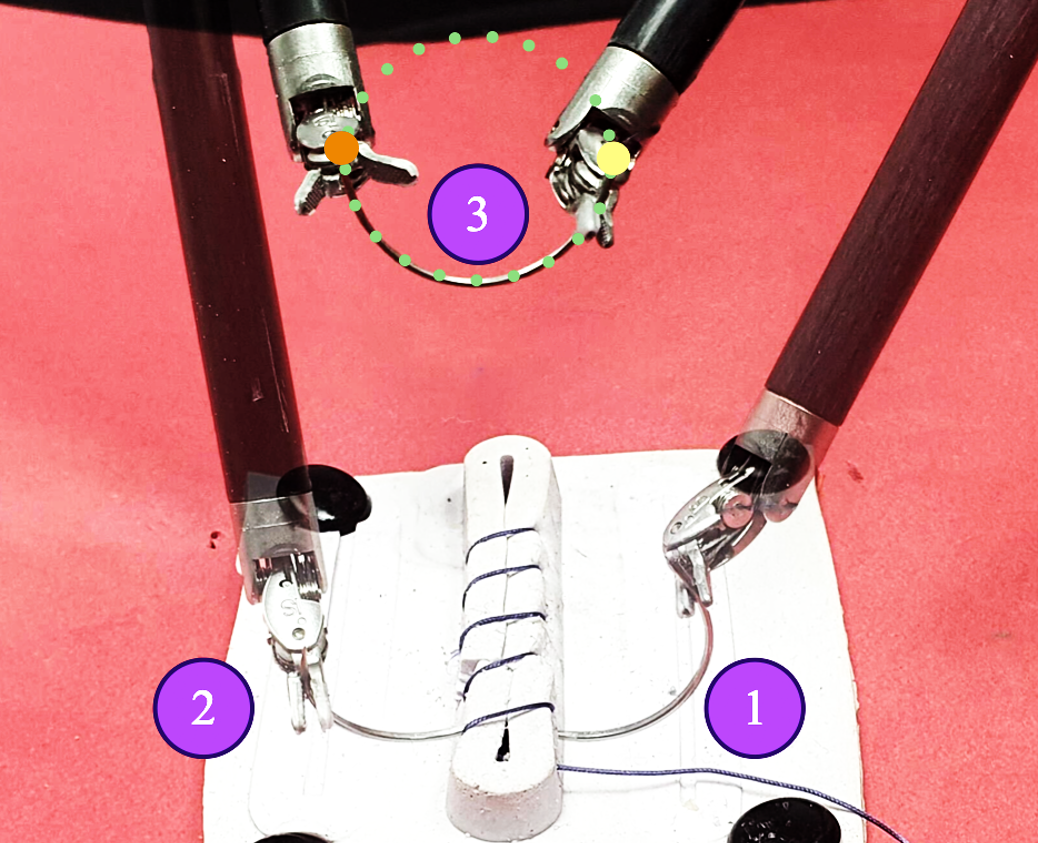}
    \vspace{-0.25in}
    \caption{\textbf{6 sutures performed by STITCH.} Step 1 shows Surgical Needle Insertion, Step 2 shows Needle Extraction, and Step 3 shows Needle Handover with Pose Correction. Detections of the needle endpoints are shown in the photo with the needle tip point shown as the orange circle and the needle swage point shown as the yellow circle. The light green circle represents the estimated needle pose.}
    \label{fig:splash}
    \vspace{-0.25in}
\end{figure}



We present STITCH, a novel pipeline that performs \textbf{S}uture \textbf{T}hrows \textbf{I}ncluding \textbf{T}hread \textbf{C}oordination and \textbf{H}andoffs. STITCH achieves "level 2" task autonomy as defined by~\textcite*{yang2017medical}, using novel perception and control techniques. The STITCH perception system combines deep learning, analytical, and sampling-based approaches, and prior knowledge of needle geometry to perform 6D needle pose estimation for closed-loop control. We also incorporate ``interactive perception" as proposed by \textcite{bohg2017} for improving needle pose estimation and correction to increase robustness to uncertainty in perception, control, and physics \cite{bohg2017}, \cite{manip}. Furthermore, we introduce visual-based thread coordination motions such as sweeping to reduce the risk of the thread tangling with the needle or itself. STITCH also includes a new approach for automated recovery from motion failures using closed loop visual feedback where we retry needle extraction and handover motions if they are unsuccessful.

This paper makes 3 main contributions:
\begin{itemize}
    \item An augmented dexterity pipeline for robot-assisted surgery capable of suture throwing, thread coordination, and needle handoff.
    \item A novel visual 6D needle pose estimation framework.
    \item Experimental results showing an average of 4.47 successful sutures with human intervention. 
\end{itemize}

\begin{figure*}[!t]
    \centering
    \includegraphics[width=0.999\linewidth]{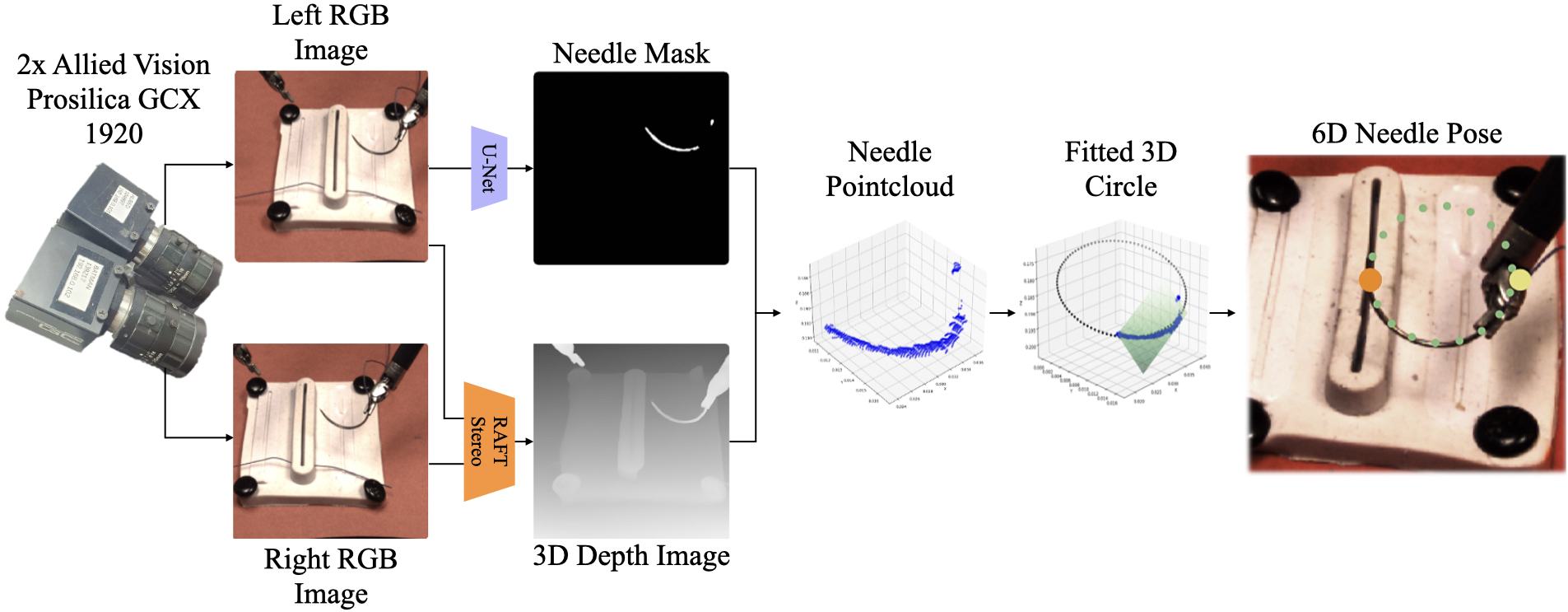}
    \vspace*{-0.3in}
    \caption{\textbf{6D Needle Pose Estimation Module.} The needle pose estimation starts with a pair of stereo left and right images. Using RAFT-Stereo, we generate a disparity image from the stereo pair \cite{lipson2021raft}. Furthermore, we segment the needle in the left image with a U-Net to create a needle mask. From there, we apply the needle mask to the disparity image, and create the corresponding needle pointcloud. Using RANSAC, we find a best-fit plane to determine the normal vector of the 3D circle representing the needle (seen in green in the Fitted 3D Circle image). Then, we project all needle inliers from the RANSAC to the plane, and use RANSAC again to find the best fit circle (seen in black in the 3D Circle Fit image with the assumed fixed radius (12 mm for all experiments). Finally, we find the two farthest points on the needle pointcloud to determine the needle endpoints (seen as orange and yellow in the 6D Needle Pose image).}
    \label{fig:needle_pose_estimation}
    \vspace*{-0.3in}
\end{figure*}
\section{Related Work}
Fully automated robot systems have been approved for hair restoration and external beam radiation \parencite{ROSE201497, kilby2010cyberknife}. 
However, all surgical procedures are performed 100\% under human surgeon teleoperation \parencite{Saeidi2022}. Some research efforts have focused on autonomously performing specific sub-tasks such as debridement \parencite{murali2015learning}, vascular shunt insertion \parencite{shunt_insertion}, and brain tumor resection \parencite{hu2018semi}. In this work, we focus on the suturing task.

\subsection{Autonomous Suturing}
Automating surgical suturing has been shown to be feasible in-vivo using IR markers, and industrial robot arms (like the KUKA LBR Med) both for the open-surgery setting \parencite{shademan2016} and the minimally invasive surgery setting \parencite{Saeidi2022} to perform in-vivo suturing on live pigs. More recent research efforts have been directed at automating surgical suturing using the da Vinci Research Kit (dVRK) \parencite{dvrk}. While performing the whole suturing task autonomously remains an open problem \parencite{suturing}, many researchers have explored the automation of sub-tasks such as suture placement \parencite{sutureplacement}, needle handover \parencite{wilcox2022learning, chiu2021, varier2020collaborative}, needle extraction \parencite{sundaresan2019automated}, needle pick-up \parencite{d2018automated}, and knot tying \parencite{thananjeyan2020safety}. Though each of these methods individually have shown high success rates (90\%), performing these methods sequentially often has much lower success rate due to the product probability intersection of all of these events. In attempts to automate the complete suturing task, some hardware simplifications have been used in the past. \textcite{schwaner2021} demonstrate impressive success rates with only a needle and no thread, avoiding the risk of the robot getting tangled in the thread. \textcite{suturing} use colored needles and a special mount for the gripper which help with needle detection and orientation. We use unmodified surgical needles and suture thread.

\subsection{Visual Servoing in Surgical Automation}
Visual servoing has been explored for surgical automation, with needle \parencite{chiu2022markerless, jiang2023} and gripper \parencite{lu2022pose} pose estimation using learned keypoint tracking models like DeepLabCut, a method that uses transfer learning to perform keypoint annotation for pose estimation by \textcite{mathis2018deeplabcut}. Another line of research focuses on learning end-to-end visual servoing policies which implicitly model the object state \parencite{levine2018learning, kalashnikov2018qt}. \textcite{paradis2021intermittent, wilcox2022learning} have proposed intermittent visual servoing in the context of peg transfer and needle handover, respectively, both of which use a visual servoing policy trained with imitation learning in lieu of classical trajectory optimization where high precision is required. We propose a novel visual needle pose estimation approach using learned image segmentation models as well as known visual features and system dynamics. The estimated object poses are used to automate the surgical suturing sub-tasks of needle insertion, extraction, and handover using analytic control methods.



\section{Problem Statement}

\subsection{Overview}
Given a sequence of suture entry and exit points, perform as many sutures as possible.

\subsection{Assumptions}
We assume known needle shape and diameter, predetermined 3D points for needle insertion and extraction, a calibrated stereo camera pair, and the transformation between the robot and camera coordinate frames. We also assume the wound is raised and orthogonal to the camera as shown in Fig. \ref{fig:splash}.



\subsection{Objectives and Evaluation Metrics}
We consider two evaluation metrics: number of successful consecutive sutures and completion time. 
We consider a suture throw to be successful if the robot is able to pass the needle through the wound, perform suture thread cinching (tightening), and return the needle to a neutral position outside the phantom with no tangles of the suture thread.

\section{Methods}
\begin{figure*}[!t]
    \vspace{0.03in}
    \centering
    \includegraphics[width=0.999\linewidth]{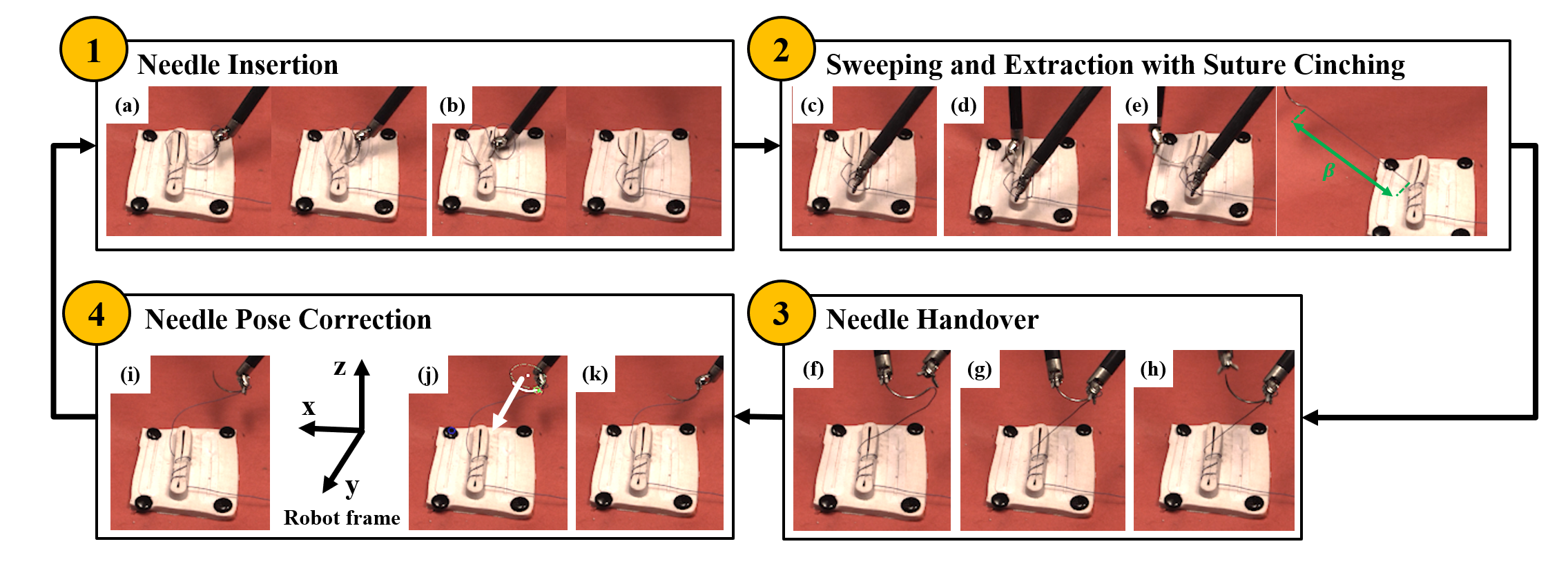}
    \vspace*{-0.25in}
    \caption{\textbf{The individual processes embedded within the STITCH motion controller} There are 4 parts to the state machine: 1. Needle insertion: (a) The right needle driver moves the needle to the initial insertion point at the proper orientation; (b) The right needle driver inserts the needle into the phantom with combined rotation and translation movements; 2. Sweeping and Needle Extraction with suture cinching: (c) The right needle driver follows the +y axis in the robot frame down the center of the wound to ``sweep'' any thread off the wound; (d) The left needle driver moves 1 centimeter behind the needle endpoint to prepare for extraction; (e) The left gripper grasps the needle and pulls it through until the length of the thread is at a desired $\beta$; 3. Needle Handover: (f) The right needle driver moves 1 centimeter behind the needle endpoint to prepare for handover; (g) The right gripper grasps the needle for handover; (h) The left gripper releases the needle; 4. Needle Pose Correction: (i) The right needle driver moves the needle to an optimal needle pose estimation region of the scene; (j) Based on the current pose of the needle, it is rotated such that the normal vector of the needle is aligned with the + y axis in robot frame; (k) The needle is rotated about the +y axis in robot frame so it is at the optimal orientation for the next insertion so the pipeline can be repeated again.}
    \label{fig:methods}
    \vspace*{-0.25in}
\end{figure*}

\AlgName achieves augmented dexterity for surgical suturing using novel perception and control methods as shown in Figs. \ref{fig:needle_pose_estimation} and \ref{fig:methods}.

\subsection{6D Needle Pose Estimation Module}\label{ssec:needle_est}
The Allied Vision Prosilica GC 1290 cameras capture a stereo pair of images with a resolution of 1280x960 at up to 33fps. With these images, we create a pointcloud of the scene using disparity images from RAFT-Stereo, a deep architecture for rectified two-view stereo that has been empirically shown to be more robust than standard stereopsis techniques and generalizes well to unseen real-world data \cite{lipson2021raft}.



To isolate the points relevant to the needle, we train a U-Net~\cite{unet}, an image-based segmentation model, to detect the needle in image space. The training data is labeled using the \textit{Labels from UV} technique proposed by \cite{luv}. The segmentation mask produced by the U-Net is used to isolate the needle from the pointcloud generated by RAFT-Stereo.

Since we also need to accurately estimate the positions of the two endpoints on the needle and the orientation of the needle itself, we propose a novel 6D needle pose estimation module as seen in Figure \ref{fig:needle_pose_estimation}. Based on Wilcox's work for needle handover (HOUSTON) \cite{wilcox2022learning}, we propose an algorithm that determines the best-fitting 3D circle. A key distinction between this work and HOUSTON is that HOUSTON obtained a pointcloud from stereo point matching while this work uses RAFT-Stereo \cite{lipson2021raft}, a roboust stereopsis neural network. STITCH begins by using RANSAC to estimate a 3D plane equation that fits the needle pointcloud. Then, we calculate the normal vector from the estimated 3D plane to ascertain the orientation of the needle. All inlier needle points are projected onto that plane. To achieve higher accuracy, we then project all inlier points on the plane to the xy-plane where we use RANSAC to estimate a 2D circle equation. Next, we derive the positions of the two endpoints of the needle on the 2D circle by finding the 2 most distant needle inliers in the pointcloud. After that, we determine the 3D circle and endpoint positions by back-projecting onto the aforementioned 3D plane estimate.



\subsection{Augmented Dexterity Suturing Motion Controller}
The surgical suturing task is composed of several distinct motions. The motion controller directs the robot motions in every state, as well as the state transitions and when they ought to be performed. Each suture is composed of the sequence of needle insertion, a thread sweeping motion to clear excess thread from the suture site, needle extraction with suture cinching, needle handover, needle pose correction, and failure recovery, as shown in Fig. \ref{fig:methods}. The needle motion inside the tissue is designed to reduce tissue damage while remaining withing the kinematic bounds of the gripper. 

\subsubsection{Needle Insertion}
The robot inserts the needle into the tissue phantom with a circular twisting motion at the specified insertion point such that the needle tip exits the tissue at the specified extraction point.

The needle is inserted in the tissue phantom in two steps to minimize strain on the tissue. Both steps are performed open loop as a large part of the needle is occluded by the tissue and the needle driver during insertion. First, the tip of the needle is pushed into the tissue phantom corresponding to the vector between the given insertion and extraction points (Fig. \ref{fig:methods}(a)). This allows for the straight tip of the needle to penetrate the tissue and exit at the needle extraction point. Second, the needle is rotated into the tissue by performing a $45^\circ$ rotation around the estimated circle normal vector (Fig. \ref{fig:methods}(b)). This motion follows the curvature of the needle as it passes through the tissue, so that it passes along the hole made by the needle tip and does not stretch or tear the tissue.

\subsubsection{Thread Sweeping}
The thread sweeping motion is designed to prevent failure during the needle extraction step (Fig. \ref{fig:methods}(c)). Failures occurring during the needle extraction process can be broadly classified into two cases: (1) the thread passes in front of the needle, occluding it and leading to detection errors; (2) when both the needle and the thread are accidentally grasped together during the re-grasping process, in spite of an accurate needle pose estimation. Pushing the thread out of the re-grasping site through a sweeping motion before the extraction step can prevent the failures described above.

We use the thread modelling method from \textcite{schorp2023selfsupervised} to track the thread for the sweeping motion. In this work, a U-Net is trained to output a segmentation mask of the thread. Running an analytic tracer on that segmentation mask for the left and right images, a 3D NURBS spline is fitted for the thread.

The sweeping motion involves opening the gripper wide and passing it over the wound towards the camera so that the thread is caught and pushed ahead of the needle. With the thread in front of the needle, entanglement risks during extraction are significantly reduced.

\subsubsection{Needle Extraction}\label{sssec:needle_aproach_and_grasp}
The purpose of the needle extraction motion is to remove the needle from the tissue phantom and to pull the thread taut to close the wound. To successfully perform multiple sutures, it is necessary to pull the thread to an appropriate length prior to the next insertion motion. The needle extraction motion is performed in a closed-loop fashion using the needle pose estimator. The frequency of needle pose estimator for feedback was measured as 0.67 Hz. The re-grasp point and the axis around which to rotate and remove the needle are also determined through visual estimation.

The needle extraction motion is composed of two main components. At the start of the routine, the left needle driver is positioned at the upper left edge of the workspace, above the tissue phantom. The point closer to the left needle driver among the two needle endpoints, determined as described in ~\ref{ssec:needle_est}, is defined as the re-grasp point. The left needle driver is moved to a point offset by 1~cm horizontally from the defined re-grasp point (Fig. \ref{fig:methods}(d). Then, the left needle driver is moved in the direction of the re-grasp point by 1.5~cm, and the gripper jaws are closed.


Once the needle has been grasped by the left needle driver, it is rotated by $80^\circ$ about the estimated needle axis to extract the needle and to minimize tissue damage (similar to the ``rotate in'') motion in the needle insertion routine. At the end of this action, only a small bit of the needle remains inside the tissue so that the needle can be fully extracted by a linear motion (Fig. \ref{fig:methods}(e)).


\subsubsection{Suture Cinching}
To ensure each suture is properly tensioned, suture cinching (tightening) is performed after needle extraction. The length of the thread that needs to be pulled for suture cinching in the extraction motion can be defined as $\beta$ = $l_{des}$ - ($i$ - 1)$\times$$l_{each}$ depending on the number of sutures (Fig. \ref{fig:methods}(e)),  where $l_{des}$ represents the desired thread length for the final suture, $i$ represents the number of sutures, and $l_{each}$ is the length of thread used for a single suture.  

\subsubsection{Needle Handover}
The goal of the needle handover motion is to transfer the extracted needle from the left needle driver to the right needle driver in preparation for the next insertion. Once the pose of the needle is estimated, we define the re-grasp point in handover as the endpoint further from the left needle driver, similar to the process in \ref{sssec:needle_aproach_and_grasp}. The right needle driver opens its jaws and moves to a point offset by 1~cm horizontally from the defined re-grasp point (Fig. \ref{fig:methods}(f)). After moving by the offset in the direction of the re-grasp point, the right needle driver closes to grasp the needle, (Fig. \ref{fig:methods}(g)) and the left needle driver opens to release it (Fig. \ref{fig:methods}(h)). The entire sequence is done in a closed-loop fashion with visual feedback tracking the pose of the needle. If the right needle driver moves into position to grasp the needle and the detected orientation of the needle remains unchanged, we pull the driver back, add a small ($<$ 0.5cm) random horizontal offset, and reattempt the grasp, up to 5 additional times before declaring the handover a failure.

\subsubsection{Needle Pose Correction}
Since the pose of needle right before insertion significantly affects the insertion, we use interactive perception to improve needle tracking so it can be actuated to the ideal pose for the subsequent insertion. We adjust the needle for the insertion using a needle pose correction algorithm, which consists of three steps. First, the needle is moved to an optimal detection location at the lower, back, right corner of the workspace for RAFT-Stereo reconstruction as shown in Fig. \ref{fig:methods}(i). The lower back right corner was chosen empirically for consistent needle pose estimates. An ablation study confirmed this choice, showing 90\% success rate in this corner for left gripper-held needles and 70\% for right gripper-held needles, compared to 50\% success in random positions. Identifying both endpoints is crucial for insertion, so obtaining pose estimates in the chosen corner maximized successful suture throws. In the next step, we sample 10 measurements of the needle normal vector, and rotate the gripper such that the normal vector of the needle is identical to the positive y axis in robot frame (Fig. \ref{fig:methods}(j)). This step constrains the discrepancy in orientation to be about the positive y-axis, which makes it more repeatable. Because the needle configuration during handover is relatively similar for each throw, the final step is a 90 degree rotation about the y-axis as seen in Fig. \ref{fig:methods}k. At this point, the STITCH pipeline is ready for the next suture throw insertion.

\begin{table*}[htbp!]
\vspace{0.25in}
\caption{\label{ablation-table}Success Metric Comparison across Ablations for 15 Trials.}
\centering
{\begin{tabular}{r|c|crrc rrcc c}
\toprule[1pt]
\multicolumn{1}{c|}{\multirow{2}{*}{}} & \textbf{Mean Sutures} & \textbf{Single-Suture} & \textbf{Three Throw} & \textbf{Full Wound} & \textbf{Mean Time} & \multicolumn{4}{c}{\textbf{Error types}} & \textbf{Mean Sutures} \\
 & \textbf{to Failure} & \textbf{Success Rate} & \textbf{Success Rate} & \textbf{Success Rate} & \textbf{per Suture} & I & E & H & T & \textbf{to Intervention} \\
\midrule[0.1pt]
\textbf{Sensing Only} & 1.40 & 51.6\% & 0.0\% & 0.0\% & 106.8 sec & 9 & 6 & 0 & 0 & – \\
\textbf{Thread Handling} & 1.80 & 55.9\% & 20.0\% & 0.0\% & 117.9 sec & 6 & 5 & 2 & 2 & – \\
\textbf{\AlgName} & 2.93 & 69.4\% & 73.3\% & 0.0\% & 159.3 sec & 8 & 5 & 0 & 2 & – \\
\textbf{\AlgName + Human} & 4.47 & 83.3\% & 100.0\% & 20.0\% & 141.9 sec & 16 & 10 & 0 & 2 & 2.25 \\
\bottomrule[1pt]
\end{tabular}}
\end{table*}

\subsubsection{Motion Failure Recovery}
The STITCH algorithm includes recovery mechanisms for both extraction motion failures and handover motion failures. For the extraction motion, the algorithm compares the positions of the needle endpoint before and after extraction. If the difference is below 2 cm, the extraction motion is retried up to 5 times. For the handover motion, the algorithm compares the normal vector of the needle before and after moving the right needle driver when both needle drivers have grasped the needle. If there is a nonzero change of the normal vector, then the handover motion is retried up to 5 times.

\section{Physical Experiments}


\subsection{Experimental Setup}
The experimental setup consists of a bi-manual dVRK robot \parencite{dvrk}, a soft tissue phantom consisting of a single wound from a 3-Dmed directional suture pad featuring parallel linear wounds, and a fixed RGB stereo camera pair. For our experiments, the stereo cameras are a pair of Allied Vision Prosilica GC 1290 industrial cameras, which each output images of size 1280x960 at 33 frames per second. The Edmund Optics lenses mounted on the sensors allow for precise adjustments to the focus and aperture based on our workspace depth and illumination. The cameras are angled at the phantom such that the full workspace of the robot is captured in the field of view. We define the workspace using a Cartesian $(x,y,z)$ coordinate system. We use violet  $\text{Polysorb}^{\text{TM}}$ surgical suture thread from Covidien. The threads are of variable length between 10 and 40 cm, with 2-0 USP Size (0.35-0.399 mm in diameter) and are attached to a GS-21 half-circular surgical needle with a radius of 12 mm.


\subsection{Baselines}
We evaluate the full STITCH algorithm and two baselines, multi-throw suturing without thread manipulation or needle pose correction ``Sensing Only") and multi-throw suturing thread management without needle pose correction (``Thread Management"). The sensing-only baseline performs the needle insertion, extraction, and handover steps, omitting the suture cinching, thread sweeping, and needle pose correction motions. The thread management baseline performs insertion, extraction (with thread sweeping), cinching, and handover while omitting the needle pose correction step. We also evaluate a human-supervised setting (``\AlgName + Human"), in which the robot can request intervention up to 2 times from a supervising surgeon. At these points, the human supervisor performs a single recovery motion to return the workspace to a safe configuration, and then immediately returns control to the robot.

\subsection{Trial Specifications and Error Classification}

Each trial produces n successive running suture throws. A suture throw is considered complete when the needle has successfully been inserted and extracted through the raised edges of the wound and the thread has been sufficiently tensioned to close the wound between the insertion and extraction points. A trial ends when the robot either enters an unrecoverable state (such as a dropped needle or tangled thread preventing further sutures) or successfully closes the wound by throwing 6 consecutive sutures. For each experimental baseline, we perform 15 trials of 1–6 throws each, and report the following metrics in Table~\ref{ablation-table}:
\begin{itemize}
    \item \textit{Mean sutures to failure}: the average number of successfully completed suture throws before the first unrecoverable error is encountered.
    \item \textit{Single-suture success rate}: the percentage of successful suture throws out of the total number of suture throw attempts.
    \item \textit{Three throw success rate}: the percentage of trials which terminate after at least three successful suture throws.
    \item \textit{Full wound success rate}: the percentage of trials which terminate in wound closure.
    \item \textit{Mean time per suture}: the average time elapsed per suture throw.
    \item \textit{Error Types}: the number of insertion (I), extraction (E), handover (H), and thread management (T) errors encountered.
    \item \textit{Mean sutures to intervention}: The average number of autonomously completed suture throws between human intervention requests.
\end{itemize}

We report trial-ending errors according to the portion of the pipeline during which the failure occurred. We use the following schema:
\begin{itemize}
    \item \textit{Insertion error}s occur when the robot fails to insert the needle through the raised edge of the wound, or the needle enters a non-wound region of the phantom, or the needle exits the wound through the top or bottom of the phantom.
    \item \textit{Extraction error}s occur when the needle remains in the wound after extraction.
    \item \textit{Handover error}s occur when the robot drops the needle during handover or enters an unrecoverable configuration during the handover process.
    \item \textit{Thread management error}s occur when the robot fails to properly cinch the suture thread to close the wound or becomes dangerously entangled in the thread.
\end{itemize}

\subsection{Experimental Results}
\begin{figure}[t]
    \centering
    \includegraphics[width=0.99\linewidth]{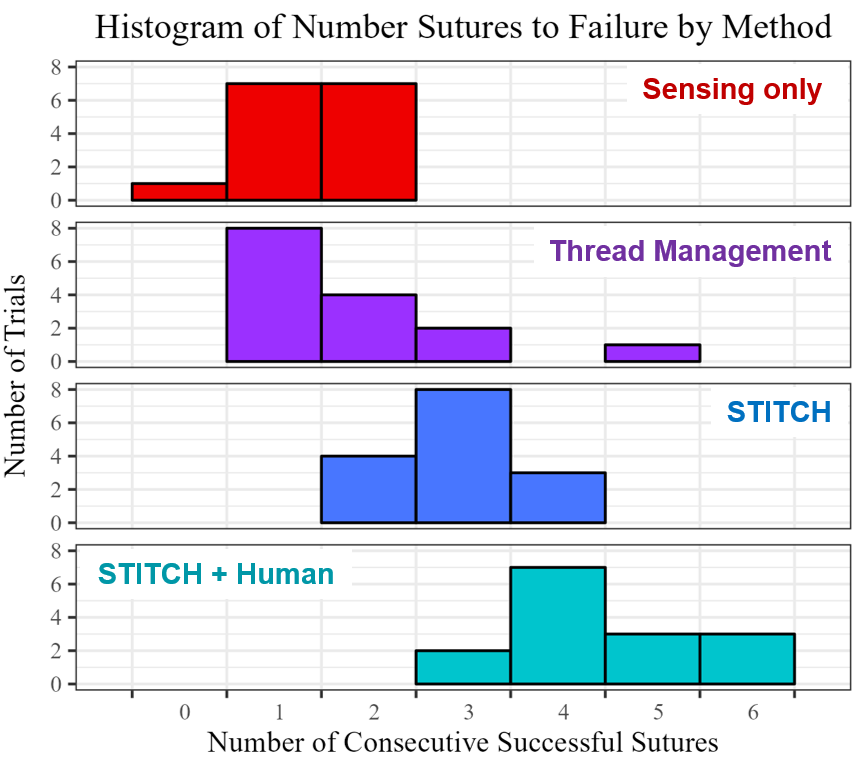}
    \vspace{-0.3in}
    \caption{\textbf{Histogram of the Number of Sutures to Failure by Method.} The results in the histogram shown above is for 15 trials per each of the four methods.}
    \label{fig:suture_histogram}
    \vspace{-0.25in}
\end{figure}

Table~\ref{ablation-table} and Fig.~\ref{fig:suture_histogram} report experimental results. Over 15 trials, \AlgName achieves an average single-suture success rate of 69.39\% and a mean sutures-to-failure of 2.93. When allowing the robot to request human intervention, the robot achieves a single-suture success rate of 83.33\%, and a mean sutures-to-failure of 4.47. Also, the perception algorithm proposed in this study was confirmed to accurately track the 6D pose of the surgical needle 42\% of the time throughout the experimental trials.

\section{Limitations \& Future Work}

There are 2 primary limitations of the STITCH pipeline.

1. The most common failure case with the STITCH pipeline involves correcting the needle orientation so it is optimal for needle insertion for the subsequent insertion. This stems from high variance needle pointclouds from RAFT-Stereo. RAFT-Stereo struggles in high disparity areas, thus we move the needle to lower, back, right corner of the workspace as seen in Fig. \ref{fig:methods}(i). At this position, the perception algorithm can reliably detect the normal vector of the needle, allowing for the first rotation correction step to align the normal vector with the positive y axis of the robot frame as seen in Fig. \ref{fig:methods}(j). However, the needle endpoint detection at this position has a higher variance than desired. This means we cannot perform the final rotation step based on needle endpoint feedback. In the future, we will investigate alternate stereo methods better tuned for small, reflective objects. With a more reliable stereopsis method and improved needle endpoint tracking, we hope to mitigate the insertion failure case by using the needle endpoint estimates to servo the needle to the optimal insertion orientation.

2. Another common failure case with the STITCH pipeline involves thread tensioning issues. Even with the thread-sweeping move, two potential thread failures are still present: 
\begin{enumerate}
    \renewcommand{\labelenumi}{\alph{enumi}.}
    \setlength{\itemsep}{6pt}
    \item \parbox[t]{\linewidth}{The sweeping move sometimes misses the thread 
    and the extraction move will grab the thread with the endpoint 
    causing system failure.}
    
    \item \parbox[t]{\linewidth}{The later sutures run out of thread and fail because not enough thread was pulled during the initial suture throw extraction.}
\end{enumerate}
In future work, we plan to introduce a suture cinching before handover where the left gripper holds the needle at the handover position such that a portion of the thread would be vertical. The other gripper would then move horizontally to push the vertical thread component to pull out additional thread and clear it from the grasping workspace.

To perform suturing in in-vivo experiments, challenges like tissue deformation, visual changes due to blood, and viewing angle variations must be addressed.

\section{Conclusion}

We present \AlgName, a continuous suturing pipeline under the Augmented Dexterity framework \cite{goldberg_augmented_2023} where surgical subtasks are performed autonomously under close human supervision. The pipeline includes a 6D needle pose estimation module using combined learned and analytical methods, and new motion primitives. Over 15 trials, we demonstrated that STITCH can autonomously perform an average of 2.93 suture throws without human intervention and 4.47 suture throws with human intervention.

\footnotesize{
    \section*{Acknowledgement}
This research was performed at the AUTOLAB at UC Berkeley in affiliation with the Berkeley AI Research (BAIR) Lab, and the CITRIS ``People and Robots" (CPAR) Initiative. We thank Gary Guthart for his support. The da Vinci Research Kit is supported by the National Science Foundation, via the National Robotics Initiative (NRI), as part of the collaborative research project ``Software Framework for Research in Semi-Autonomous Teleoperation" between The Johns Hopkins University (IIS1637789), Worcester Polytechnic Institute (IIS 1637759), and the University of Washington (IIS 1637444).)
}

\printbibliography


\end{document}